%% file: main.tex
\documentclass[letterpaper]{article}
\usepackage{uai2019}
\usepackage[margin=1in]{geometry}

\usepackage{times}

\usepackage{hyperref}
\usepackage{url}
\input{sections/headers.tex}

\title{Learning to learn generative programs with Memoised Wake-Sleep}

\author{
{\bf Luke B. Hewitt}~~~~~
{\bf Tuan Anh Le}~~~~~
{\bf Joshua B. Tenenbaum}\\
\texttt{\{lbh,tuananh,jbt\}@mit.edu}\\
Department of Brain and Cognitive Sciences, MIT
}

\begin{document}

\maketitle

\input{sections/body.tex}

\bibliography{bib}
\bibliographystyle{iclr2020_conference}

\clearpage
\appendix
\onecolumn
{
\toptitlebar \centering
{\Large\bf Learning to learn generative programs with Memoised Wake-Sleep\\Supplementary Material}
\bottomtitlebar
}

\input{sections/supplement.tex}

\end{document}

%% file: sections/headers.tex
\usepackage{natbib}
\usepackage{capt-of}
\usepackage{lipsum}
\usepackage{stfloats}

\usepackage{wrapfig}
\usepackage{booktabs}
\usepackage{subfig}
\usepackage[ruled]{algorithm2e}
\usepackage{algorithmic}
\usepackage{multirow}
\usepackage{amsmath}
\usepackage{amsfonts}
\usepackage{wrapfig}
\usepackage{stackengine}
\usepackage{xcolor}
\usepackage{graphicx}
\usepackage{mdframed}
\newmdenv[topline=false,rightline=false,bottomline=false]{leftborder}

\usepackage{amssymb}
\usepackage{amsthm}
\usepackage{caption}

\input{sections/math_commands.tex}

\usepackage{floatrow}
\newfloatcommand{capbtabbox}{table}[][\FBwidth]

\newcommand{\LongName}{Memoised Wake-Sleep}
\newcommand{\Name}{MWS}
\newcommand{\NameRemember}{{\Name }-\textit{replay}}%
\newcommand{\NameSleep}{{\Name-\textit{fantasy}}}%
\newcommand{\RWSSleep}{RWS-\textit{sleep}}%
\newcommand{\CSVConcepts}{String-Concepts}

\newcommand{\gray}[1]{\textcolor{gray}{#1}}
\newcommand{\Regex}[1]{\texttt{\textbf{#1}}}

\newcommand{\RegexPriorRow}[2]{
#1 & #2 \\
}

\newcommand{\Str}[1]{\textsf{#1}}
\newcommand{\comma}{,\hspace{0.25cm}}
\newcommand{\bs}{\textrm{\textbackslash}}

\newcommand{\Ra}{{\color{red} .}}
\newcommand{\Rd}{{\color{red} \bs d}}%
\newcommand{\Rl}{{\color{red} \bs l}}%
\newcommand{\Ru}{{\color{red} \bs u}}%
\newcommand{\Rw}{{\color{red} \bs w}}%
\newcommand{\Rs}{{\color{red} \bs s}}%
\newcommand{\RK}{{\color{blue}*}}
\newcommand{\RP}{{\color{blue}+}}
\newcommand{\RA}{{\color{blue}|}}
\newcommand{\RM}{{\color{blue}?}}

\newcommand{\kl}{\textrm{D}_\textrm{KL}}
\newcommand{\EE}{\mathop{\mathbb{E}}}

%% file: sections/math_commands.tex
\usepackage{amsmath,amsfonts,bm}

\def\eqref#1{equation~\ref{#1}}

\def\1{\bm{1}}

\def\eps{{\epsilon}}

\DeclareMathAlphabet{\mathsfit}{\encodingdefault}{\sfdefault}{m}{sl}
\SetMathAlphabet{\mathsfit}{bold}{\encodingdefault}{\sfdefault}{bx}{n}

\newcommand{\KL}{D_{\mathrm{KL}}}

%% file: sections/body.tex
\begin{figure*}[b]
\vspace{0.8em}
\begin{minipage}[b]{\textwidth}
\hspace{-0.5 em}
\begin{minipage}[c]{0.25\textwidth}
\includegraphics[width=0.95\textwidth]{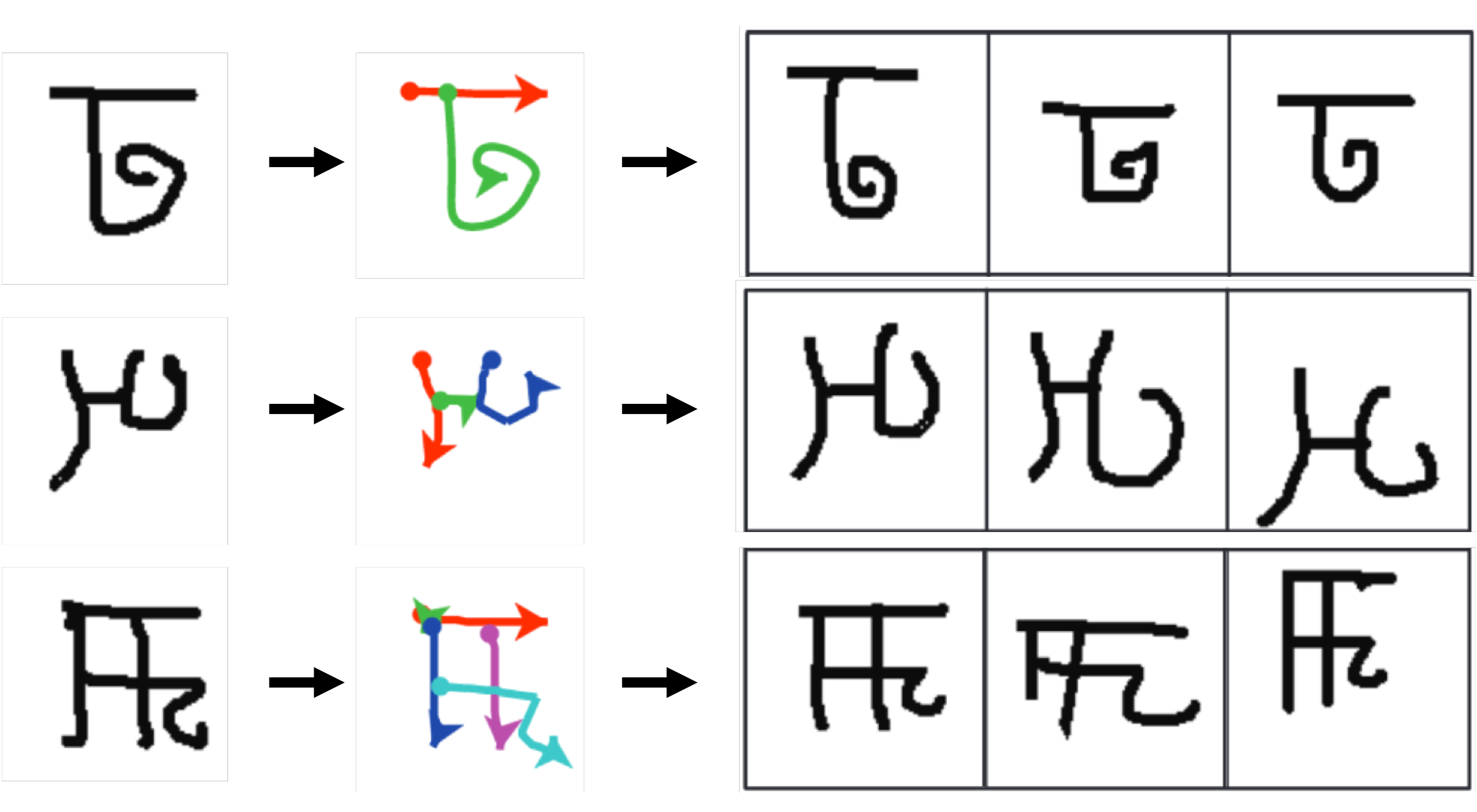}
\end{minipage}
\begin{minipage}[c]{0.49\textwidth}
\centering
\textsf{
\tiny
\def\arraystretch{1.3}
\begin{tabular}{|r|r|r|r|r|
} \hline
21.8\%&B,J,U&img25.png&\$4,050
&Jul 10, 2012\\
7.5\%&B,D,E&img19.png&\$4,000
&Nov 16, 2004\\
-10.1\%&A&img42.jpg&\$340
&Nov 1, 2000\\
22.4\%&A,B,J,U&img18.png&\$102
&Jun 9, 2016\\
-4.0\%&B,E&img43.jpg&\$1,200
&Nov 1, 1999\\
\hline
\end{tabular}
}
\label{string-concepts}
\end{minipage}
\begin{minipage}[c]{0.20\textwidth}
\includegraphics[width=0.95\textwidth]{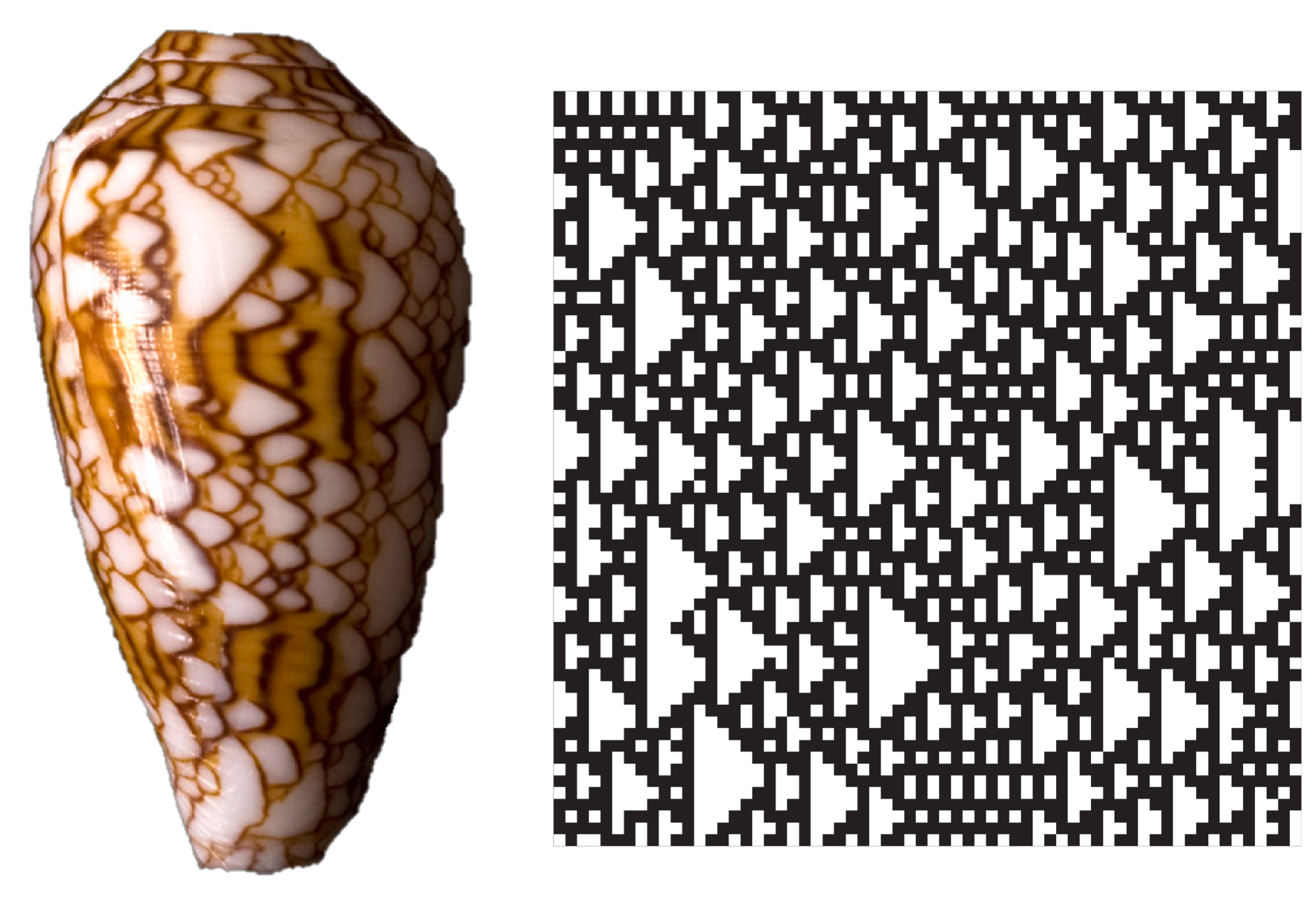}
\end{minipage}

\vspace*{-0.5em}
\captionof{figure}{\label{fig1}Finding compositional structure can enable generalisation from minimal data. People can infer strokes in a character to produce novel variations \citep{lake2015human} or learn string concepts from a few examples (our dataset). Seashells patterns can be described and extrapolated using simple cellular automaton rules \citep{wolfram2002new}.\vspace{0.5em}\\{
\small{\textit{Proceedings of the 36\textsuperscript{th} Conference on Uncertainty in Artificial
 \vspace{-0.2em}\\Intelligence (UAI)}, PMLR volume 124, 2020.}}
}
\label{fig:concepts}
\end{minipage}
\end{figure*}

\begin{abstract}
    We study a class of neuro-symbolic generative models in which neural networks are used both for inference and as priors over symbolic, data-generating programs. As generative models, these programs capture compositional structures in a naturally explainable form.
    To tackle the challenge of performing program induction as an `inner-loop' to learning, we propose the Memoised Wake-Sleep (MWS) algorithm, which extends Wake Sleep by explicitly storing and reusing the best programs discovered by the inference network throughout training.
    We use MWS to learn accurate, explainable models in three challenging domains: stroke-based character modelling, cellular automata, and few-shot learning in a novel dataset of real-world string concepts. 
    \vspace{-1em}
\end{abstract}

\section{INTRODUCTION}
\vspace{-0.6em}
From the phonemes that make up a word to the
nested goals and subgoals that make up a plan, 
many of our models of the world rely on symbolic structures such as categories, objects, and composition.
Such explicit representations are desirable not only for interpretability, but also because models that use them are often highly flexible and robust. For example, in spreadsheet editing, FlashFill uses program inference to allow specification of batch operations by example \citep{gulwani2015inductive}. In character recognition, the stroke-based model of \citet{lake2015human} remains state-of-the-art at both few-shot classification and generation, despite competition from a variety of neural models \citep{lake2019omniglot}.

In this work we focus on structured generative modelling: we aim to find symbolic \textit{generative programs} to describe a set of observations, while also learning a prior over these programs and fitting any continuous model parameters. For example, we model handwritten characters by composing a sequence of strokes, drawn from a finite bank of stroke types which is itself learned.

Unfortunately, learning such models from scratch is a substantial challenge.
A major barrier is the difficulty of search: discovering a latent program for any given observation is challenging due to the size of the space and sparsity of solutions. Furthermore, this inference must be revised at every iteration of learning.

We build on the \textit{Helmholtz Machine} \citep{dayan1995helmholtz}, a longstanding approach to learning in which two models are trained together: a generative model $p(z)p(x|z)$ learns a joint distribution over latent and observed variables, while a recognition network $q(z|x)$ performs fast inference. This approach, including more recent variants such as VAEs \citep{kingma2013auto}, is well-suited to learning \textit{neural} generative models because, as noted by \citet{hinton1995wake} \textit{``the algorithm adapts the generative weights so as to make $p(-|x)$ close to $q(-|x)$"}. That is, when $p(x|z)$ is a neural network, the semantics of the latent space are highly unconstrained, and so can be learned to aid fast recognition.

\begin{figure*}[t]
    \centering
    \includegraphics[width=0.99\textwidth]{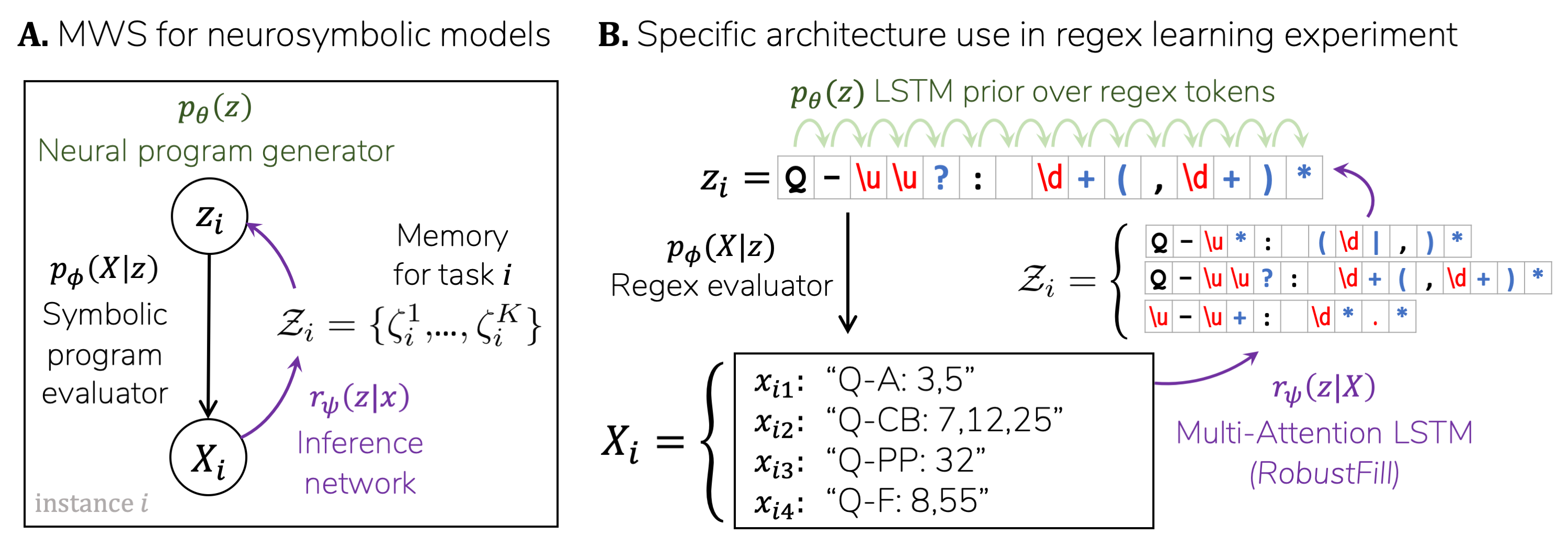}
    \caption{
    \textbf{A.} In {\LongName }, we train $p$ using samples from a finite set $\mathcal{Z}_i$, containing the best $K$ programs found for $X_i$. We use a recognition network $r_\psi(z|X)$ to propose updates to this set. \textbf{B.} For our regex model, each $X_i$ is a set of strings generated by a latent regex $z_i$. As a recognition network, we use a program synthesis LSTM (RobustFill, \citet{devlin2017robustfill}) to propose regexes for each set $X_i$.
    \vspace{-1em}
    }
    \label{fig:main-diagram}
\end{figure*}

Unlike such purely neural generative models, the models we consider have a more constrained and interpretable latent space.
We take $z$ to be a sequence of discrete tokens representing a data-generating program. Our goal is to learn a prior $p_\theta(z)$ over programs (which may be a neural network such as an LSTM), alongside parameters of a symbolic program evaluator $p_\phi(x|z)$ and a program recognition network $r_\psi(z|x)$.  Figure~\ref{fig:main-diagram} describes such a model for string concepts, in which $z$ is a regular expression and $p_\phi$ is a symbolic regex parser.

In this setting, training a network for fast and accurate inference is ambitious: even state-of-the-art program induction networks often must guess-and-check thousands of candidates. This is impractical for Helmholtz machine algorithms, which require inference to be repeated at each gradient step.
We propose a novel algorithm, \textit{\LongName } ({\Name}), which is better suited for the structured domains we study. Rather than restart inference afresh at each iteration, {\Name } maintains a finite set $\mathcal{Z}_i =
\{\zeta_i^1, \zeta_i^2, \ldots, \zeta_i^M\}$ containing the best programs discovered by the recognition network for instance $x_i$, remembering and reusing these to train the generative model in future iterations.

Our contributions in this paper are as follows. We first outline the {\Name } algorithm, and prove that it optimises a variational bound on the data likelihood, where $\mathcal{Z}_i$ is the support of a finite variational distribution $Q_i(z_i)$.
We then illustrate {\Name } with a simple nonparametric Gaussian mixture model and evaluate on
three structure-learning domains (Fig.~\ref{fig:concepts}), finding that it greatly outperforms more computationally intensive approaches of Reweighted Wake Sleep \citep{bornschein2014reweighted} and VIMCO \citep{mnih2016variational} while often providing an significant speedup. We also develop a novel \textit{\CSVConcepts} dataset, collected from publicly available spreadsheets for our evaluation. This dataset contains 1500 few-shot learning problems, each involving a real-world string concept (such as \textit{date} or \textit{email}) to be inferred from a small set of examples.

\section{BACKGROUND}

The Helmholtz Machine \citep{dayan1995helmholtz} is a framework for learning generative models, in which a recognition network is used to provide fast inference during training.
Formally, suppose we wish to learn a generative model $p_{\theta\phi}(z,x) = p_\theta(z)p_\phi(x|z)$, which is a joint distribution over latents $z$ and observations $x$, and a recognition network $q_\psi(z| x)$, which approximates the posterior over latents given observations. The marginal likelihood of each observation is bounded by:
\begin{align}
&\log p_{\theta\phi}(x) 
\ge \log p_{\theta\phi}(x) - \kl[q_\psi(z|x)||p_{\theta\phi}(z|x)] \label{eq:helmholtz-bound1}\\
&= \EE_{z \sim q_\psi(z| x)} \log p_{\phi}(x|z) - \kl[q_\psi(z|x)||p_\theta(z)] \label{eq:helmholtz-bound2}
\end{align}
where $\kl[q_\psi(z|x)||p_{\theta\phi}(z|x)]$ is the KL divergence from the true posterior $p_{\theta\phi}(z|x)$ to the approximate posterior $q_\psi(z|x)$.
Learning is typically framed as maximisation of this \textit{evidence lower bound} (\textit{ELBO}) by training the recognition network and generative model together.

Gradient-based maximisation of this objective with respect to $\theta$ and $\phi$ is straightforward: an unbiased gradient estimate for Eq.~\ref{eq:helmholtz-bound2} can be created by taking a single sample $z \sim q_\psi(z|x)$ each gradient step. However, maximising Eq.~\ref{eq:helmholtz-bound2} with respect to $\psi$ is more challenging and two main approaches exist:

\textbf{VAE.} We may update $\psi$ also using an unbiased estimate of Eq.~\ref{eq:helmholtz-bound2}, sampling $z \sim q_\psi(z|x)$. However, if $z$ is a discrete symbolic expression, then estimating the gradient requires the REINFORCE estimator \citep{williams1992simple,mnih2014neural}. Despite advances in control-variate techniques, this estimator often suffers from high variance, which may lead to impractically slow training.\footnotemark{}

\textbf{Wake-Sleep.} Instead of using the KL term that appears in Eq.~\ref{eq:helmholtz-bound1}, we may update $q_\psi$ \textit{approximately} by minimising the reversed KL divergence $\KL[p_{\theta\phi}(z|x)||q_\psi(z|x)]$. In practice, this means updating $q_\psi(z|x)$ at each iteration using data sampled from the model's prior $z,x~\sim~p_{\theta\phi}(z,x)$. This yields an algorithm which is not in general convergent, yet still often performs competitively if $z$ is discrete.

Beyond the optimisation difficulties that come with discrete latent variable modelling, a further challenge arises when the recognition model $q_\psi(z|x)$ is simply incapable of matching the true posterior $p_{\theta\phi}(z|x)$ accurately. This is common even in deep generative models, which can flexibly adapt their latent representation. To address this, the above approaches may be extended by taking multiple samples $z_1,\ldots,z_K$ from the recognition model at each training iteration, then using importance weighting to estimate the true posterior. For VAEs, this yields the \textit{Importance Weighted Autoencoder} (IWAEs,  \citet{burda2015importance}), and is often applied to discrete variables using multiple samples for variance reduction (VIMCO, \citet{mnih2016variational}). For Wake-sleep, it yields \textit{Reweighted Wake-Sleep} (RWS, \citet{bornschein2014reweighted}), in which the recognition model may be trained either by the generative model (\textit{RWS-sleep}) or the importance-weighted posterior (\textit{RWS-wake}).

\footnotetext{Many approaches for training discrete VAEs are inapplicable to the models we study here. \citet{rolfe2016discrete} constructs architectures for which discrete variables can be marginalised out, while relaxation techniques \citep{jang2016categorical} approximate discrete variables continuously to produce low-variance gradient estimators or control variates \citep{tucker2017rebar,grathwohl2017backpropagation}. These methods are intractable for the compositional symbolic models we consider, as they require exponentially many path evaluations (\citet{le2018revisiting}). The \textrm{EC}$^2$ algorithm \citep{ellis2018learning} enables inference in such compositional models, but does not learn model parameters.}

\section{MEMOISED WAKE-SLEEP}

\vspace{-1em}
Our goal is learning and inference in rich neurosymbolic models such as that shown in Figure~\ref{fig:main-diagram}, for which all parameters are continuous, and the latent variables are symbolic programs.
These models pose a challenge for Helmholtz machines: given the strong constraints on $z$, it is common that only a small set of latent programs can well-explain any given observation $x_i$, and these may be difficult for $q_\psi$ to recognise quickly and reliably. The importance-weighted methods described above (RWS, VIMCO) may therefore require evaluating very many samples $z_k\sim q_\psi(z|x)$ per iteration to train $p_{\theta\phi}$. This is computationally wasteful, as it amounts to re-solving the same hard search problems repeatedly.

\vspace{-0.4em}
We propose an alternative approach which actively utilises the sparsity of good solutions in $p_{\theta\phi}(z|x_i)$ to its advantage. In the {\LongName } algorithm we do not discard the result of inference after each training step.
Instead, for each observation $x_i$ we introduce a memory $Q_i$ containing a set of the best distinct historical samples from the recognition model. Formally, we take $Q_i$ to be a variational distribution over $z_i$, which has finite support $\mathcal{Z}_i = \{\zeta_i^1, \ldots, \zeta_i^M\}$
and probabilities $Q_i(\zeta) \propto p_\theta(\zeta, x_i)$.
In the box below, we prove two statements which suggest a simple algorithm for updating $Q_i$, which maximises the ELBO (Eq.~\ref{eq:helmholtz-bound1}) by minimising\vspace{-0.5em}:
\begin{align*}
L = D_{KL}[ Q_i(z) || p(z|x_i) ]
= \sum_{\zeta \in \mathcal{Z}_i}  Q_i(\zeta) \log \frac{Q_i(\zeta)}{ p(\zeta, x_i)}  + C
\end{align*}

\fbox{
\begin{minipage}[h]{0.47\textwidth}
\begin{align*}
    \textrm{Let } Q_i(\zeta) = \sum_{m=1..M} w_i^m
    \delta_{\zeta_i^m}(\zeta),
    \textrm{\hspace{0.5em} with} \sum_{m}w_i^m = 1.
\end{align*}

\textbf{Claim 1}
Fixing the support of $Q_i$ to \hspace{-0.3em} $\mathcal{Z}_i = \{\zeta_i^1, \ldots, \zeta_i^M\}$\\ the optimal weights are given by $w_i^m \propto p(\zeta_i^m, x_i)$.
\vspace{0.5em}

\textit{Proof.}
At optimality, there can be no pair $(m, m')$ for which $L$ is reduced by moving probability mass from $Q(\zeta_i^{m})$ to $Q(\zeta_i^{m'})$.
We therefore solve by setting $dL/dw_i^{m} = dL/dw_i^{m'}$. Hence,
\vspace{-0.5em}\begin{align*}
    w_i^{m}/p(\zeta_i^{m}, x_i) &= w_i^{m'}/p(\zeta_i^{m'}, x_i), \textrm{ for all } (m, m').
\end{align*}

\textbf{Claim 2}
Fixing the weights $w_i$, we decrease $L$ if we replace any $\zeta_i^m$ with a new value $\zeta'$, such that $\zeta' \notin \mathcal{Z}_i$ and $p(\zeta', x) > p(\zeta_i^m, x_i)$.
\vspace{0.5em}

\textit{Proof.}
Rewriting the loss as
\vspace{-0.3em}
\begin{align*}
L = \sum_m w_i^m (\log w_i^m - \log p(\zeta_i^m, x_i)) + C
\end{align*}
we see that the only dependence on $\zeta_i^m$ is through $[-w_i^m \log p(\zeta_i^m, x_i)$].
The update therefore satisfies:
\begin{align*}
    \Delta L = w_i^k \log p(\zeta_i^k, x_i) - w_i^k \log p(\zeta', x_i) < 0.
\end{align*}

\end{minipage}
}

Repeated application of claims 2 and 1 yields an intuitive algorithm for optimising $Q_i$.
Every iteration, we sample a set of new programs $z^1, \dotsc, z^R$ from a recognition network, which we call $r_\psi(z| x_i)$, and compare those to the programs already in memory ($\mathcal{Z}_i$).
We then update the memory to contain the best $M$ unique elements from either the sampled programs or the existing memory elements, ranked by $p_{\theta\phi}(\cdot, x_i)$.
We then resample a program $z_Q \sim Q_i$ from memory to train $p_{\theta\phi}(z,x)$.

\begin{algorithm*}[t]
\input{sections/algo.tex}
\label{alg:trainingalgorithm}
\end{algorithm*}

To train the recognition network, we propose two variants of our algorithm. In {\NameSleep }, we train $r_\psi$ on $z,x$ pairs sampled directly from the generative model $p_\theta(z)p_\phi(x|z)$, as in the sleep phase of the wake-sleep algorithm. In {\NameRemember }, we train $r_\psi$ on the same pair $x_i,z_i$ that was sampled from memory $Q_i$ to train $p$ (analogous to RWS-\textit{wake}, \citet{bornschein2014reweighted}). In practice we find that the latter performs well, and is significantly faster as it requires no additional sampling. In this paper, we therefore refer to {\NameRemember } and RWS\textit{-wake} as simply MWS and RWS, but include additional results for {\NameSleep } and RWS-\textit{sleep} in the appendix.

\begin{figure}[h]
\centering
\begin{minipage}{\textwidth}
\centering
    \includegraphics[width=0.9\textwidth]{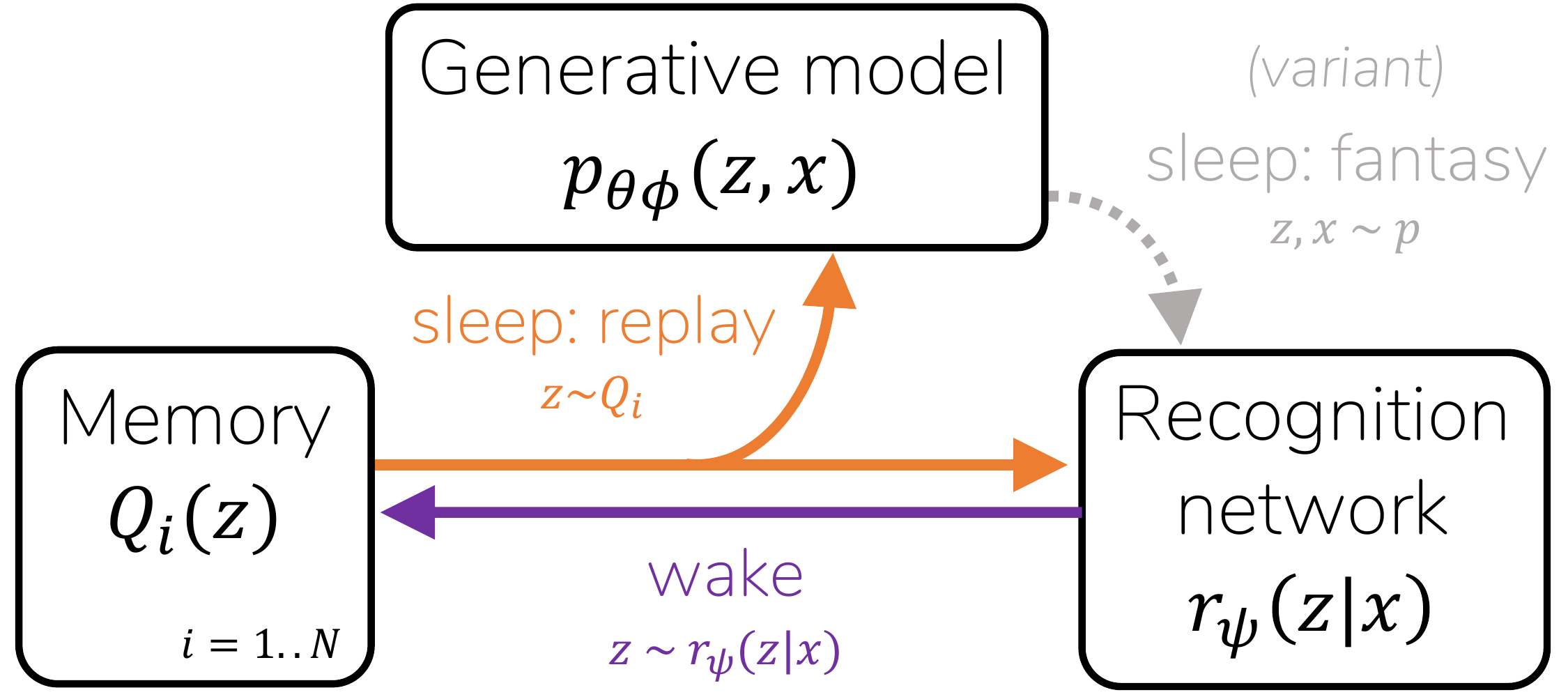}
\end{minipage}
\caption{
MWS extends Wake-Sleep with a separate memory $Q_i$ for each observation $x_i$. This memory is a discrete distribution defined over a finite set $\mathcal{Z}_i$. Each phase of {\Name} uses a sample from one model to update another:
during \textit{wake}, the recognition network samples a program $z \sim r(z|x_i)$ and, if $p(z, x_i)$ is large, $z$ is stored in memory $Q_i$.
During \textit{sleep:replay}, $z$ is sampled from $Q_i$ and used to train $p$ and $r$. Alternatively, $r$ may be trained by sampling from $p$ (in a \textit{sleep:fantasy} phase).\vspace{-1em}}
\label{fig:whattrainswhat}
\end{figure}

The three phases of the algorithm (\textit{wake}, \textit{sleep:replay} and  \textit{sleep:fantasy}) are summarised in Figure \ref{fig:whattrainswhat}, and the full algorithm is provided above.
Unlike RWS and VIMCO, the memory usage of {\LongName } contains a term linear in the dataset size $O(MN)$, due to maintaining a separate set $\mathcal{Z}_i$ of programs for each instance $i$. However, in practice this is typically negligible compared the reduction of memory required for training the recognition network: MWS can achieve strong performance with many fewer recognition samples ($R$) per iteration. 

The memory size, $M$, may be chosen to trade off accuracy and efficiency, with $M=1$ corresponding only to MAP inference, and $M\rightarrow \infty$ approaching full Bayesian inference over $z$. For modest values of M, achieving a small variational gap $D_{KL}(Q_i||p(z|x_i))$ relies on sparsity in the true posterior, as $Q_i(z)$ will converge to the best $M$-support posterior approximation of $p(z|x_i)$.\footnote{See appendix for more discussion of MWS limiting behaviour, including an empirical study of sparsity in $p(z|x)$.}

In this paper, we use default values of $M=R$, and define $K=M+R$ when presenting results. This means MWS is matched to baseline algorithms on the number of the $p_{\theta\phi}(z, x)$ evaluations per iteration ($K$), but requires half as many recognition model evaluations ($R$).

\input{sections/gmm.tex}

\input{sections/chars.tex}

\input{sections/regex.tex}

\input{sections/cellular.tex}

\section{DISCUSSION}
\vspace{-0.8em}
This work sits at the intersection of program learning and neural generative modelling.
From a dataset, we aim to infer a latent \textit{generative program} to describe each observation, while simultaneously learning a neural prior over programs and any additional model parameters.

To tackle the challenge of performing program induction repeatedly during learning, we train our models by a novel algorithm called Memoised Wake-Sleep (MWS), and find that this improves the quality of learning across all domains we study. MWS builds upon existing Helmholtz machine algorithms by maintaining a memory of programs seen so far during training, which reduces the need for effective amortized inference.
We optimise a variational bound previously proposed by \citet{saeedi2017variational}, extending their algorithm to include parameter learning for a recognition network and for the generative model.
Algorithmically, our approach is also similar to memory-based reinforcement learning methods~\citep{abolafia2018neural,liang2018memory} which maintain a queue of best action-traces found during training.

In general, MWS can be applied to any models for which all parameters are continuous, and all latent variables are discrete.
However, we particularly advocate its use for the class of `programmatic' generative models we study%
, due to the difficulties of sparse inference that they often present. If learning can be made tractable in such models, they have the potential to greatly improve generalisation in many domains, discovering compositional structure that is rich and understandable.

\vspace{-1.1em}
\paragraph{Acknowledgements.} This work was supported by AFOSR award FA9550-18-S-0003 and the MIT-IBM Watson AI Lab.
\newpage

%% file: sections/algo.tex
\def\stackalignment{r}
  \hspace{1em} initialize $\theta, \phi, \psi$ \hfill \textit{Parameters for prior $p_\theta(z)$, evaluator $p_\phi(x|z)$, recognition network $r_\psi(z|x)$}\\
  
  \hspace{1em} initialize $\mathcal{Z}_i = \{\zeta_i^m\}_{m=1..M}$ for $i=1..N$\hfill \textit{For each instance $i$, $\mathcal{Z}_i$ is a set of $M$ distinct programs}\\

  \hspace{1em} \textbf{repeat}\\
    \hspace{2em}\textrm{draw instance} $(i, x_i)$ \textrm{from dataset}\\
    
\hspace{2em}\textit{Wake}\hspace{4.5em}$\begin{cases}
z^1, \ldots, z^R \sim r_\psi(-|x_i)\\
\mathcal{\tilde Z} \gets \textrm{unique}(\zeta_i^1,\ldots,\zeta_i^M,z^1,\ldots,z^R)\\
\mathcal{Z}_i \gets \textrm{best } M \textrm{ values in } \mathcal{\tilde Z}\textrm{, sorted by } p_{\theta\phi}(\cdot, x_i)\\
\end{cases}$\hfill \textit{\stackanchor{Update memory with}{\stackanchor{samples from recognition}{network}}}

\hspace{2em}\textit{Sleep: replay}\hspace{1.3em}$\begin{cases}
z_Q \sim \mathcal{Z}_i \textrm{ with probability } Q_i(\zeta_i^k)\propto p_{\theta\phi}(\zeta_i^k,x_i)\\
(\theta, \phi) \gets (\theta, \phi) + \lambda \nabla_{\theta\phi} \log p_{\theta\phi}(z_Q, x_i)\\
\psi \gets \psi + \lambda \nabla_\psi \log r_\psi(z_Q|x_i) \textbf{ *}\\
\end{cases}$\hfill
\hspace{-1em}\textit{
\stackanchor{Train generative \& recognition}{models on sample from memory}
}\\

  \hspace{2em}\textit{Sleep: fantasy}\textbf{ *}\hspace{0.26em}$\begin{cases}
z_p, x_p \sim p_{\theta\phi}\\
\psi \gets \psi + \lambda \nabla_\psi \log r_\psi(z_p| x_p)
\end{cases}$\hfill
\textit{\Big(Optional variant: \stackanchor{Train recognition network on}{ sample from generative model } \Big)}\\
\texttt{\\}

  \hspace{1em} \textbf{until} convergence
  \caption{{\Name} training procedure (batching omitted for notational simplicity). We refer to the learning rate as $\lambda$, and to the joint probability $p_\theta(z)p_\phi(x|z)$ as simply $p_{\theta\phi}(z,x)$. In the optional algorithmic variant, MWS-\textit{fantasy}, the recognition model is instead training using a sample from $p_{\theta\phi}$ (\textit{i.e.} only one $\psi$-step marked * is required).}

%% file: sections/gmm.tex
\clearpage

\begin{figure*}[!t]
    \centering
    \includegraphics[width=1\textwidth]{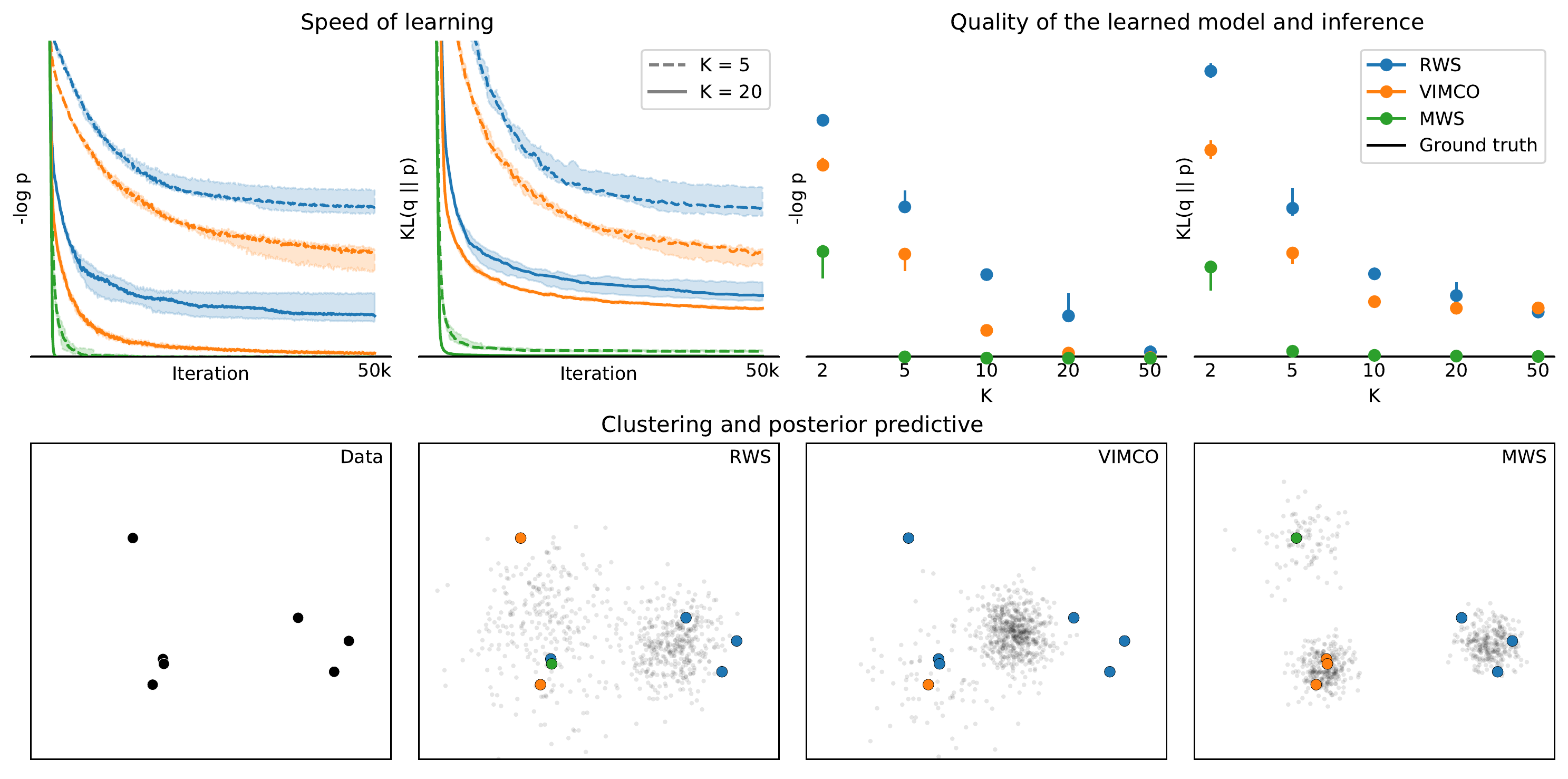}
    \caption{Quantitatively \emph{(top)} MWS outperforms RWS and VIMCO in terms of both speed of convergence and quality of learned model and inference. MWS learns near-perfect model and inference parameters with only $K = 5$ particles.
    Qualitatively \emph{(bottom)}, the neural amortized inference of RWS and VIMCO fails to cluster accurately. The model compensates by increasing the within-cluster variance $\Sigma$, as seen by the more spread out posterior predictive (gray).
    }
    \label{fig:gmm}
\end{figure*}

\section{EXPERIMENTS}

\subsection{GAUSSIAN MIXTURE MODEL}

We first validate the MWS algorithm for learning and inference in a simple nonparametric Gaussian mixture model, for which we can evaluate model performance exactly. In this model, the latent variable $z$ corresponds to a clustering of datapoints. MWS is therefore well-suited because the latent space is discrete and exponentially large in the number of data points, while the true posterior is highly peaked on few clusterings of the data.

We generated a synthetic dataset of 100 mini-datasets, with each comprising $J=7$ two-dimensional data points, as illustrated in the bottom-left of Fig.~\ref{fig:gmm}.
The latent variable for each dataset $x := x_{1:J}$ is a sequence of cluster assignments $z := z_{1:J}$ and a mean vector $\mu_c$ for each cluster $c$.
The only learnable model parameter is $\Theta \in \mathbb R^{2 \times 2}$ which parameterizes the cluster covariance.

We use a Chinese restaurant process (CRP) prior for $z$ in order to break the permutation invariance of clustering and to avoid fixing the number of clusters.
The full generative model is therefore given by:
\begin{align*}
    z_{1:J} &\sim \mathrm{CRP}(\alpha)\\
    \mu_c &\sim \mathcal N(0, I), &\textrm{for } c = 1..\infty \\
    x_j &\sim \mathcal N(\mu_{z_j}, \Sigma=\Theta\Theta^T),&\textrm{for } j=1..J
\end{align*}

\newpage
In this model, the cluster means may be marginalized out analytically, allowing us to exactly calculate $p_\theta(z, x)$ during learning. For the recognition model $r_\psi(z | x)$ we use a feedforward neural network with one hidden layer and a $\tanh$ activation whose output logits are masked to enforce valid sequences under the CRP prior.

We train the model using Adam with default hyperparameters for $50$k iterations for $K \in \{2, 5, 10, 20, 50\}$, and evaluate model quality using the average negative log marginal likelihood.
For inference quality, we evaluate $\textsc{KL}(q(z | x) || p(z | x))$ where $p(z | x)$ is the posterior under the true data-generating model and $q(z | x)$ is the memory-induced posterior approximation for MWS and the importance-weighted recognition-based approximation for RWS and VIMCO. In Figure~\ref{fig:gmm} (top), we show medians and inter-quartile ranges of the marginal likelihood and the \textsc{KL} from $10$ runs of training. 

With a moderate number of particles $K$, both RWS and VIMCO algorithms fail to cluster the data accurately using the recognition network. By contrast MWS can maintain a persistent, high-quality approximation to the true posterior for each mini-dataset. This discrepancy of inference quality is shown by the sample clustering in Figure~\ref{fig:gmm} (bottom). In turn, the use of inaccurate inference during training causes RWS and VIMCO converge to a model $p_\theta$ with poorer marginal likelihood.

\clearpage

%% file: sections/chars.tex
\clearpage

\begin{figure*}[t]
    \centering
    \includegraphics[width=1\textwidth]{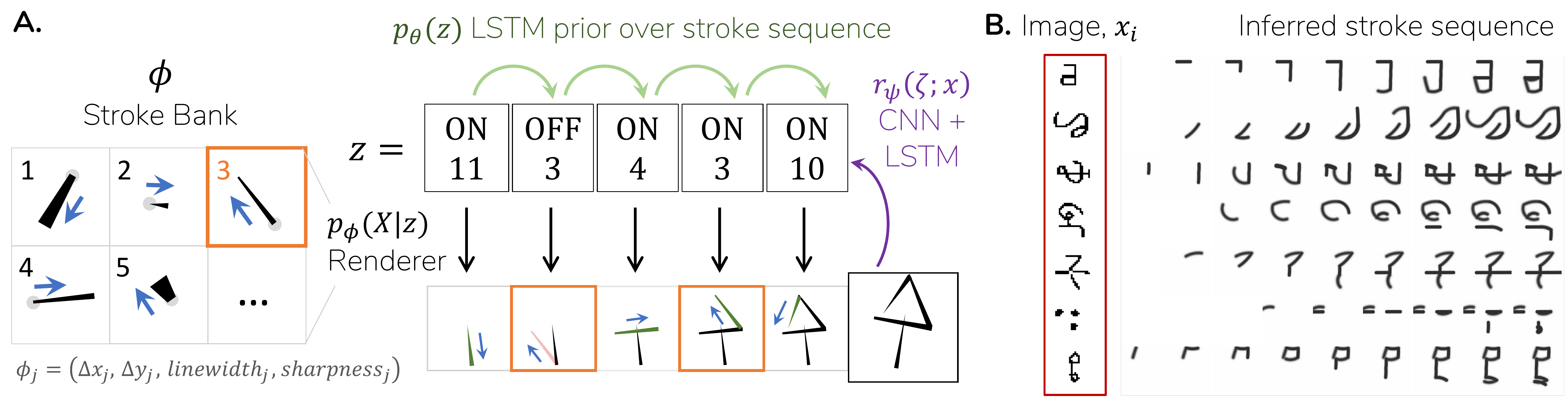}
    \caption{\textbf{A.} Schematic for our stroke-based model of handwritten characters. $p_\theta(z)$ is a prior over tokens sequences $z$, where each token $j$ points to a particular stroke from a finite set. The available strokes are learned parameters of the renderer $p_\phi(x|z)$, varying in length, direction, thickness and sharpness, curvature. Strokes chosen in $z$ are placed end to end on a canvas (optionally marked as OFF for pen movement without drawing). \textbf{B.}~Reconstructions of Omniglot characters. In each row we sample a program $z \sim Q_i$, and visualise the canvas of the renderer $p_\phi(x|z)$. 
    }
    \label{fig:characters}
\end{figure*}

\subsection{DRAWING HANDWRITTEN CHARACTERS}
\label{section-chars}
Next, we build a generative model of handwritten characters using an explicit stroke-based renderer. Drawing inspiration from \citet{lake2015human}, our model contains a finite bank of learnable stroke types, varying in parameters such as length, thickness, direction and curvature. Each latent variable $z$ is a sequence of integers which index into this bank. For generation, the renderer $p_\phi(x | z)$ places the corresponding strokes sequentially onto a canvas, which is differentiable with respect to the stroke parameters $\phi$ (Fig.~\ref{fig:characters}A). To calculate $p_\phi(x|z)$, we use this canvas to provide Bernoulli pixel probabilities, marginalising across a set of affine transforms in order to allow programs to be position invariant.
The prior $p_\theta(z)$ and the recognition network $r_\psi(z | x)$ are LSTMs and recognition network additionally takes as input an image embedding given by a convolutional network.

In Fig.~\ref{fig:characters}B, we visualise the stroke sequences inferred by our model after training on a random subset of characters from the Omniglot dataset (approximately 10 characters per alphabet across 50 alphabets). For each character, we sample a program $z$ from the memory $Q$ of the MWS algorithm, and visualise the render canvas at each step of $z$. We find that our model is able to accurately reconstruct a wide variety of characters, and does so using a natural sequencing of pen strokes.
In Table~\ref{tab:omniglot_nll}, we compare the performance of baseline algorithms applied to the same model. We find that MWS is able to learn effectively with very few particles, yet continues to outperform alternative algorithms at learning even up to $K=40$.

\begin{table}[t]
    \centering
        \begin{tabular}{l|cccccc}
            & $K=3$ & $5$ & $10$ & $20$ & $40$ \\ %
            \hline

            RWS & 0.363 & 0.348 & 0.333 & 0.324 & 0.322 \\
            {VIMCO} & 0.361 & 0.333 & 0.326 & 0.318 & 0.319 \\ 
            MWS & \textbf{0.311} & \textbf{0.305} & \textbf{0.321} & \textbf{0.310} & \textbf{0.316}
        \end{tabular}
    \caption{Marginal NLL (bits/pixel, avg. of 3 runs).}
    \label{tab:omniglot_nll}
    \vspace{-1em}
\end{table}

Our approach combines the strengths of previous work on modelling handwritten characters.
Like \citet{ganin2018synthesizing}, we learn only from raw images, aided by a neural recognition model.
However, like \citet{lake2015human}, we use a symbolic representation of characters:
our model uses a limited symbolic vocabulary of 64 strokes rather than allowing the model to produce a free-form stroke at each time step, and we restrict $z$ to a maximum of only 10 strokes. This provides an inductive bias that should encourage reuse of strokes across characters,
potentially allowing our model to make richer generalisations.

\begin{figure}[h]
    \centering
    \includegraphics[width=\textwidth]{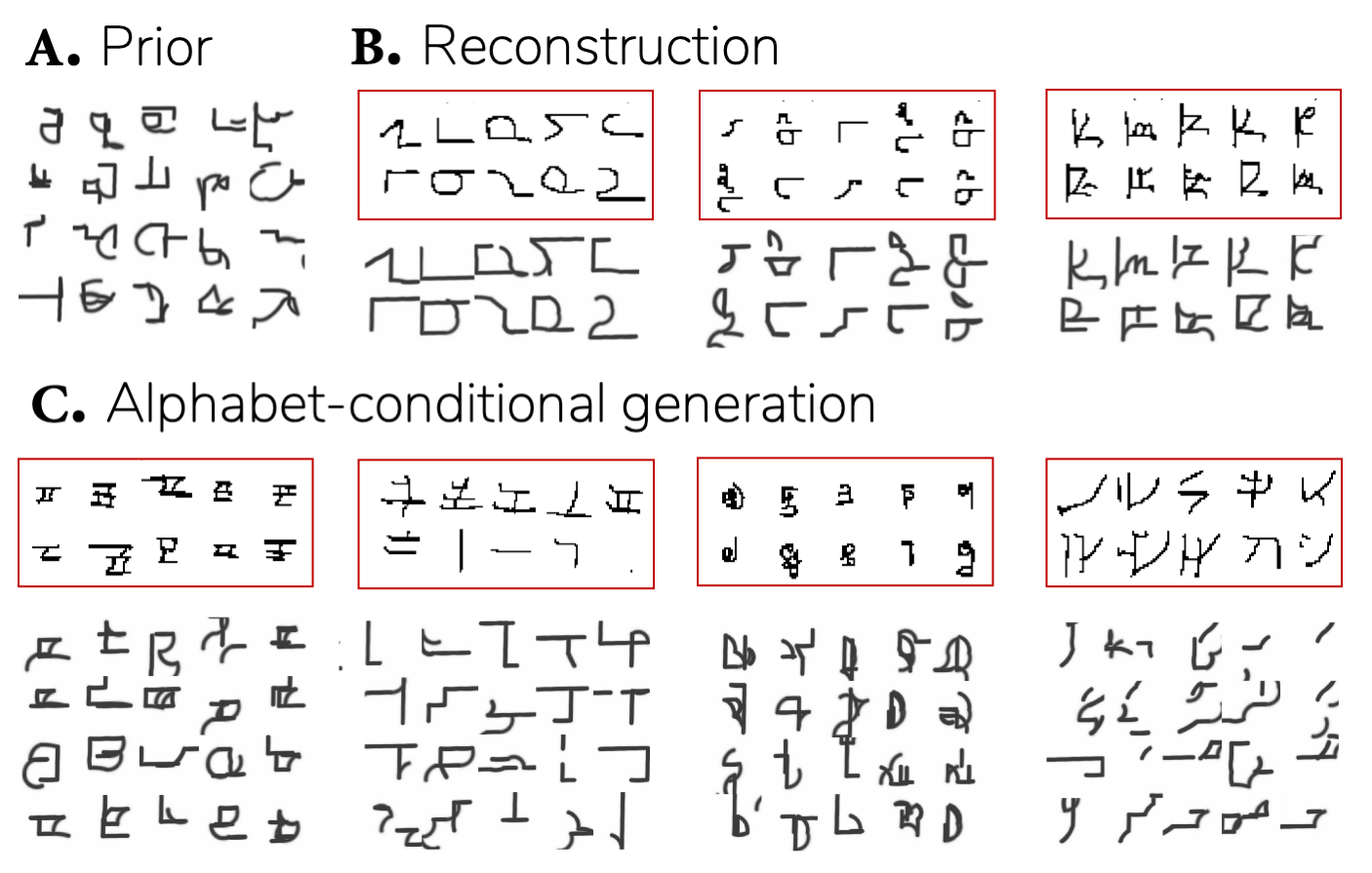}
    \caption{\textbf{A.} Samples from unconditional model. \textbf{B.} and \textbf{C} Samples from the alphabet-conditional model, for both instance reconstruction and novel character generation.}
    \label{fig:alphabets2}
\end{figure}

To illustrate this, we extend our model by conditioning the prior and the recognition model on the alphabet label which we provide during training. Given the ten characters from an alphabet (red), this conditioned-model can generate novel samples which somewhat capture its high-level style by reusing common patterns (Fig.~\ref{fig:alphabets2}C).

%% file: sections/regex.tex
\begin{figure*}[t]
\begin{minipage}{\textwidth}
\begin{minipage}[c]{0.32\textwidth}%
\scriptsize
\centering
\def\arraystretch{1.2}
\begin{tabular}{|l|}
\hline
    \multicolumn{1}{|c|}{\textbf{Character classes}}\\
    \Regex{\Ra} $\rightarrow$ any character \gray{($\boldsymbol{\phi}_\Ra$)}\\
    \Regex{\Rw} $\rightarrow$ alphanumeric character \gray{($\boldsymbol{\phi}_\Rw$)}\\
    \Regex{\Rd} $\rightarrow$ digit \gray{($\boldsymbol{\phi}_\Rd$)}\\
    \Regex{\Ru} $\rightarrow$ uppercase character \gray{($\boldsymbol{\phi}_\Ru$)}\\
    \Regex{\Rl} $\rightarrow$ lowercase character \gray{($\boldsymbol{\phi}_\Rl$)}\\
    \Regex{\Rs} $\rightarrow$ whitespace character \gray{($\boldsymbol{\phi}_\Rs$)}\\
    \multicolumn{1}{|c|}{\rule{0pt}{1.3em}$\boldsymbol{\phi}$ contains specific probabilities}\\
    \multicolumn{1}{|c|}{for each allowed character}\\
\hline
\end{tabular}
\end{minipage}
\begin{minipage}[c]{0.35\textwidth}\vspace{-0.3em}
\scriptsize
\centering
\def\arraystretch{1.2}
\begin{tabular}{|ll|}
\hline
    \multicolumn{2}{|c|}{\textbf{Operators}}\\
    \multicolumn{2}{|l|}{\textit{Optional subexpression}}\\
    \quad\Regex{E\RM} &$\rightarrow$ \Regex{E} \gray{($\phi_\RM$)} $\mid$ $\epsilon$ \gray{($1-\phi_\RM$)}\\
    \multicolumn{2}{|l|}{\rule{0pt}{1.3em}\textit{Repetition}}\\
    \quad\Regex{E\RK} &$\rightarrow$ \Regex{E\RP} \gray{($\phi_\RK$)} $\mid$ $\epsilon$ \gray{($1-\phi_\RK$)}\\
    \quad\Regex{E\RP} &$\rightarrow$ \Regex{EE\RK}\\
    \multicolumn{2}{|l|}{\rule{0pt}{1.3em}\textit{Either/or}}\\
    \quad\Regex{E$_1$\RA E$_2$} &$\rightarrow$ \Regex{E$_1$} \gray{($\phi_\RA$)} $\mid$ \Regex{E$_2$} \gray{($1-\phi_\RA$)}\\
    \multicolumn{2}{|r|}{\rule{0pt}{1.3em}$\phi$ contains production probabilities}\\
\hline
\end{tabular}
\end{minipage}
\begin{minipage}[c]{0.32\textwidth}
\centering
    \includegraphics[width=\textwidth]{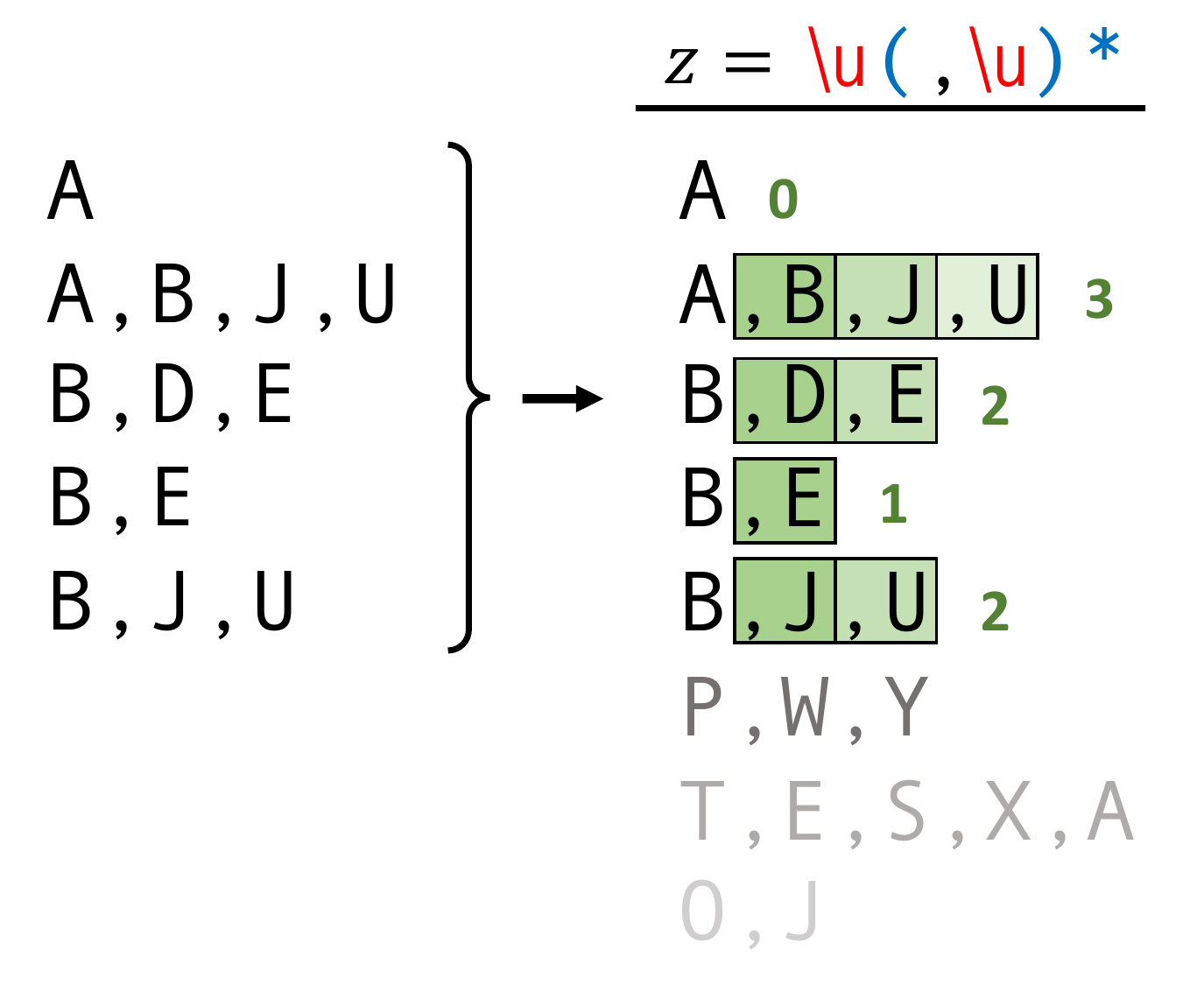}
\end{minipage}
\captionof{figure}{Character classes (\textit{left}) and operators (\textit{center}) included in our probabilistic regex model. Parameters $\phi$ determine the probability of a regex producing any given string $p_\phi(x|z)$, which can be calculated exactly by dynamic programming.
\textit{Right}: Given five example strings, the model finds a plausible regex explanation $z=\Regex{\Ru(,\Ru)\RK}$ which can be used to generates novel instances.
The inferred repeating subexpression \Regex{(,\Ru)\RK} is highlighted in green.
}
\label{fig:regexparse}
\end{minipage}
\end{figure*}

\subsection{STRUCTURED TEXT CONCEPTS}

We next apply {\Name} at learning short text concepts, such as \textit{date} or \textit{email address}, from a few example strings. This task is of interest because such concepts often have a highly compositional and interpretable structure, while each is itself a fairly rich generative model which can be applied to generate new strings. 

For this task we created a new dataset comprising 1500 concepts, each with 5 training strings and 5 test strings, collected by random sampling of spreadsheet columns crawled from public GitHub repositories. The data was filtered to remove columns that contain only numbers, English words longer than 5 characters, or common names (see Figure~\ref{fig1} and Appendix for dataset samples).

We aim to model this dataset by inferring a regular expression (regex) $z$ for each concept. This is a convenient choice because regexes can naturally express compositional relationships, and can be evaluated efficiently on any given string. Specifically, we consider \textit{probabilistic} regexes: programs which \textit{generate} strings according to a distribution, and for which the probability of any given string $p(x|z)$ can be calculated exactly and efficiently.

The full model we develop for this domain is shown in Fig.~\ref{fig:main-diagram}. We use an LSTM prior over regexes $p_\theta(z)$, a program-synthesis LSTM network $r_\psi(\zeta|x)$ to infer regexes from strings (RobustFill, \citet{devlin2017robustfill}), and a symbolic regex evaluator $p_\phi(x|z)$. The prior and recognition networks output a sequence of regex tokens, including characters (4, 7, \textit{etc.}), character classes (\Regex{\Rd } for digit, \Regex{\Ru } for uppercase, \textit{etc.}), operators (\Regex{\RK } for repetition, \textit{etc.}) and brackets. In the regex evaluator $p_\phi(x|z)$, learnable parameters determine the assignment of probability to strings: for example, when \Regex{\RK} appears, the number of repeats is geometrically distributed with parameter $\phi_\RK$. The full set of parameters $\phi$ is shown in Figure.~\ref{fig:regexparse}.

\begin{figure*}[t]
    \centering
    \includegraphics[width=0.97\textwidth]{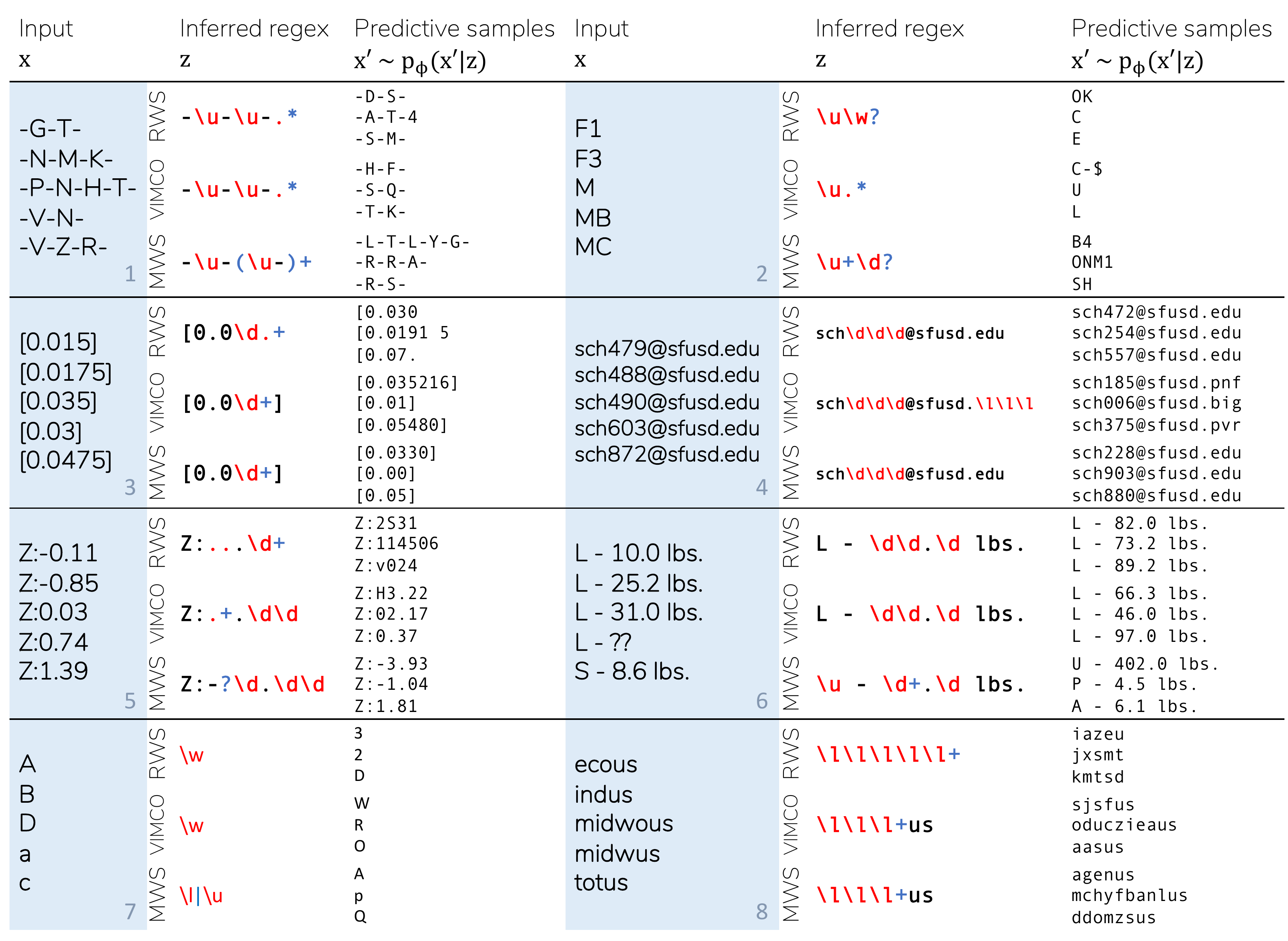}
    \caption{Inferred regexes and posterior predictive samples from models trained on the \textit{\CSVConcepts} dataset. Posterior samples are taken from $Q_i$ in MWS, and from $q(z|x_i)$ with K-importance sampling in RWS and VIMCO.}
    \label{fig:regexes}
\end{figure*}

We first present the results from training this model using the MWS algorithm, with $K=10$. Table~\ref{regexprior} shows prior samples generated by the learned model. For each row, we draw a new concept from the LSTM prior $z \sim p_\theta(z)$, and then generate several instances using the symbolic regex evaluator, sampling $x \sim p_\phi(x|z)$. This demonstrates qualitatively that our model generalises meaningfully from the training concepts: the invented concepts are highly plausible, containing motifs such as \# preceding a string of digits, or \% as a final character. The model's prior has learned high level structures that are common to many concepts in the training data, but can compose these structures in novel ways.

In Table~\ref{tab:regex_marginal_nll} we quantitatively evaluate {\Name }, RWS and VIMCO algorithms for the same neurosymbolic model architecture (additional results with MWS-\textit{fantasy} and RWS-\textit{sleep} \hspace{0.1em}variants \hspace{0.1em}are \hspace{0.1em}provided \hspace{0.1em}in \hspace{0.1em}the \hspace{0.1em}appendix). \hspace{0.1em} We

\vspace{2em}
\begin{minipage}[h]{0.49\textwidth}
\centering
\input{sections/regexprior.tex}
  \captionof{table}{Novel concepts sampled from the MWS model. In each row we sample a regex $z$ from the learned prior, then generate examples from this regex.}
  \label{regexprior}
\end{minipage}

estimate the true marginal likelihood of all models on held out test data using importance sampling. %
{\Name } not only exceeds the performance of baselines algorithms for large values of $K$, but also achieves strong performance with only $K=2$ particles. This allows a model to be fit accurately at a very significant reduction in computational cost, requiring $5\times$ fewer evaluations of $p_{\theta\phi}(z,x)$ per iteration. Such efficiency is particularly valuable for domains in which likelihood evaluation is costly, such as those requiring parsing or enumeration when scoring observations under programs.

Figure~\ref{fig:regexes} shows qualitatively the inferences made by our model. Across a diverse set of concepts, the model learns programs with significant compositional structure. In many cases, this allows it to make strong generalisations from very few examples.

Furthermore, comparison across algorithms demonstrates that these more complex expressions are challenging for the recognition network to infer reliably. For example, while RWS and VIMCO are typically able to discover template-based programs (e.g. Figure~\ref{fig:regexes} concept 4), MWS is the only algorithm to utilise operators such as alternation (\Regex{\RA}) or bracketed subexpressions for any concepts in dataset (e.g. concepts 7 and 1).

\vspace{1em}
{
\centering
\begin{minipage}[b]{0.45\textwidth}
\centering
    \begin{tabular}{l|cccc}
            & $K=2$ & $3$ & $5$ & $10$\\
            \hline
            
            {RWS} & 87.1 & 86.5 & 85.2 & 85.0\\
            {VIMCO} & 97.5 & 89.1 & 84.8 & 83.5\\
            {\Name } & \textbf{84.1} & \textbf{83.1} & \textbf{82.8} & \textbf{82.5}
            \vspace{-0.8em}
        \end{tabular}
        \label{tab:regexnll}
    \captionof{table}{Marginal NLL (nats, avg. of 3 runs)}
    \label{tab:regex_marginal_nll}

\end{minipage}
}

%% file: sections/regexprior.tex
{\def\arraystretch{1.6 }
\small
\begin{tabular}{|ll|}
\hline
 \multicolumn{1}{|l}{\small \textbf{Prior} $z \sim p_\theta$} & \multicolumn{1}{l|}{ \small \textbf{Generated $x \sim p_\phi(x|z)$}}\\
 \hline

\RegexPriorRow
{\Regex{c\Rs\Rd\Ra\Rd\RP}}
{\Str{c 0.6}{\comma}\Str{c 4.4}{\comma}\Str{c 6.0}
}

\RegexPriorRow
{\Regex{\Rw\Rd\Rd\Rd\Rd-\Rd\Rd}}
{\Str{56144-73}{\comma}\Str{60140-63}
}

\RegexPriorRow
{\Regex{\$\Rd00}}
{\Str{\$600}\comma\Str{\$300}\comma\Str{\$000}
}

\RegexPriorRow
{\Regex{\Rl\Rl\Rd}}
{\Str{hc8}{\comma}\Str{ft5}{\comma}\Str{vs9}
}

\RegexPriorRow
{\Regex{\#\Rd\Rd\Ru\Ru}}
{\Str{\#57EP}\comma\Str{\#11UW}\comma\Str{\#26KR}
}

\RegexPriorRow
{\Regex{\Ru0\Rd\Rd\Rd\Rd\Rd\Rd}}
{\Str{B0522234}\comma\Str{M0142810}
}

\RegexPriorRow
{\Regex{\Ru\Ru\Ru\Rd\Ra\Rs\Rd0\%}}
{\Str{TAP0. 70\%}\comma\Str{THR6. 50\%}
}

\RegexPriorRow
{\Regex{R0<\Rd\RP}}
{\Str{R0\texttt{<}3}\comma\Str{R0\texttt{<}9}\comma\Str{R0\texttt{<}80}
}

\RegexPriorRow	
{\Regex{\Ru\RP.}}
{\Str{EA.}\comma\Str{SD.}\comma\Str{CSB.}
}

\hline
\end{tabular}}

%% file: sections/cellular.tex
\begin{figure*}[t]

\begin{minipage}{0.55\textwidth}
\includegraphics[width=\textwidth]{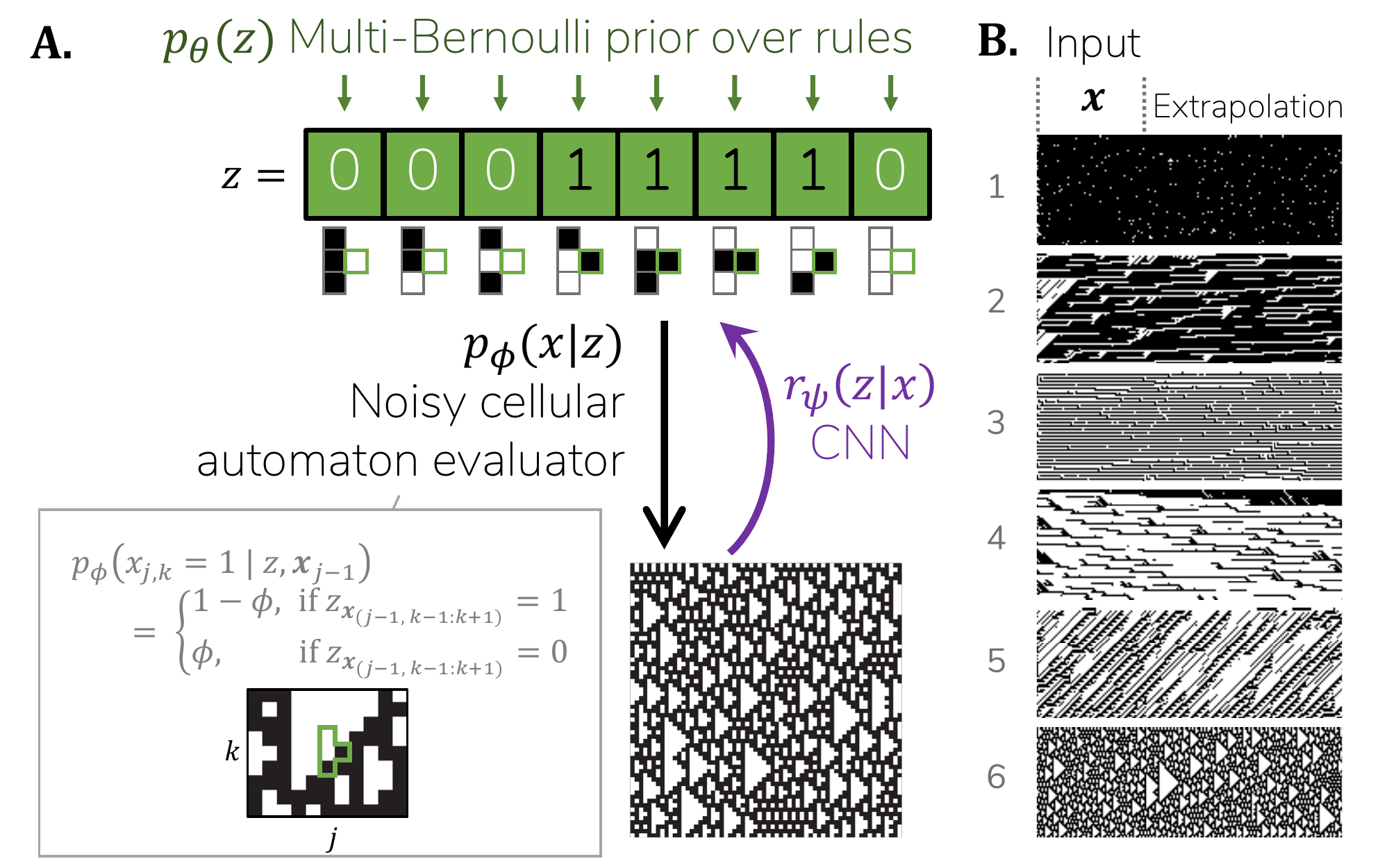}
\end{minipage}
\begin{minipage}{0.4\textwidth}
\centering
    \begin{tabular}{l|cccc}
            & $K=2$ & $3$ & $5$ & $10$\\
            \hline
            
            {RWS} & 0.74 & 0.51 & 0.17 & 0.09\\
            {VIMCO} & 1.31 & 1.05 & 0.66 & 0.37\\
            {\Name } & \textbf{0.01} & \textbf{0.01} & \textbf{0.02} & \textbf{0.02}
            \vspace{-0.8em}
        \end{tabular}
        
    \vspace{1.5em}
    Distance from $\phi$ to true model $\eps$ (easy task)
    
    \vspace{1em}
    
        \begin{tabular}{l|cccc}
            & $K=2$ & $3$ & $5$ & $10$\\
            \hline
            
            RWS & 10.27 & 5.55 & 3.44 & 2.43 \\
            {VIMCO} & 5.98 & 3.39 & 2.15 & 1.26\\
            {\Name } & \textbf{1.24} & \textbf{1.15} & \textbf{0.90} & \textbf{0.75}
        \end{tabular}

    \vspace{1em}
    Distance from $\phi$ to true model $\eps$ (hard task)
    
\end{minipage}

    \caption{
    \textbf{A.}~In our model, $z$ describes a CA in canonical binary form (depicted is Rule 30, \citet{wolfram2002new}). Images are generated from left to right, with each pixel stochastically conditioned on its three left-neighbours. \textbf{B.} MWS is able to infer the CA rule for each image, and learns a global noise parameter $\phi$, which we then use to extrapolate the images. The model accurately matches the true generative noise, as is most visually apparent in row 1.}

    \label{fig:cellular-results}
\end{figure*}

\subsection{NOISY CELLULAR AUTOMATA}
Finally, to demonstrate the use of {\Name } in estimating meaningful parameters, we consider the domain of cellular automata (CA). These processes have been studied by \citet{wolfram2002new} and are often cited as a model of seashell pigmentation. We consider \textit{noisy, elementary} automata: binary image-generating processes in which each pixel (cell) is sampled stochastically given its immediate neighbours in the previous column (left to right).

For this domain, we build a dataset of $64 \times 64$ binary images generated by cellular automata. For each image we sample a ``rule'', $z$, which determines the value of each pixel given the configuration of its immediate 3 neighbours in the previous column. Each rule is represented canonically as a binary vector of length $2^3 = 8$ (see Figure~\ref{fig:cellular-results}). To generate an image $x$, the leftmost column $\mathbf{x}_1$ is sampled uniformly at random, and then subsequent columns are determined by applying the rule $z$ to each pixel $x_{j,k}$, with corruption probability $\eps$ (fixed to $2\%$):
\vspace{-0.5em}\begin{align*}
p(x_{j,k}=1 | z, \mathbf{x}_{j-1}) =\begin{cases}
1-\eps, &\textrm{if } z_{(\mathbf{x}_{j-1,k-1:k+1})}=1\\
\eps, &\textrm{otherwise}.
\end{cases}
\end{align*}

\vspace{-0.7em}We then build a generative model which matches this process, but where the rules are latent variables and the noise $\eps$ is replaced by a learnable parameter $\phi$. We aim to perform joint learning and inference in this model, and use $\phi$ to estimate the true $\eps$. This estimation a significant challenge, because any inaccurate inferences of $z$ will cause the model will overestimate the noise $\eps$ to compensate. As a stringent test of the algorithm, we also evaluate on a harder domain of \textit{nonelementary} automata, with cells depending on five neighbours (so $z$ has length $32$).

In Figure \ref{fig:cellular-results}B, we visualise the automata inferred by {\Name} by extrapolating images from the dataset. It is visually apparent that the model accurately captures the generative process for each image~$x_i$, including both the rule~$z_i$ and noise~$\eps$. This is confirmed quantitatively: for both the easy and hard task variants, {\Name} is able to estimate the true generative process parameter with significantly greater accuracy than alternative algorithms.

%% file: sections/supplement.tex
\section{MEMORY IN PERCEPTUAL RECOGNITION}

\begin{figure}[ht]
    \centering
    \includegraphics[width=\textwidth]{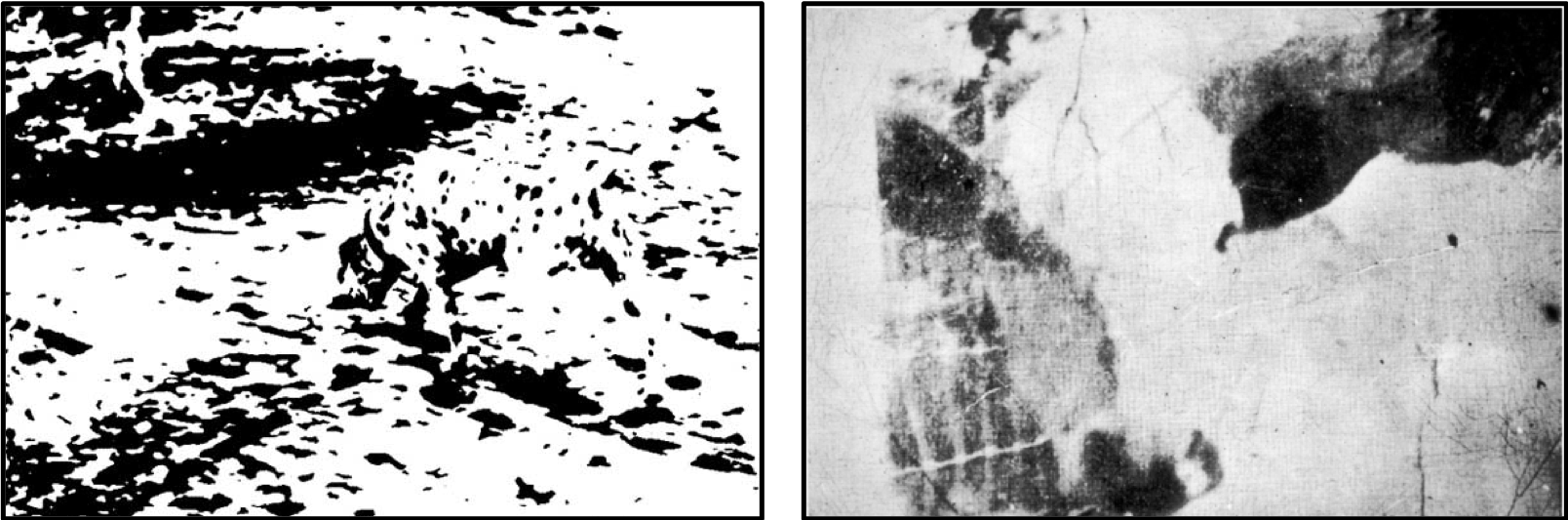}
    \caption{\textbf{What is depicted in these images?} Viewers typically find it difficult to see the subject of these images on first presentation, suggesting that it involves solving a very difficult search problem. However, after initial recognition, this structure is immediately apparent on all future viewings, as though the result of previous inference has been stored for future reuse. Answer in footnote}
\end{figure}

\footnotetext{Left: a dalmatian sniffs the ground, \citep{gregory1970intelligent}. Right: a cow looks towards the camera, \citep{kmd1951puzzle}.}

\section{MWS-FANTASY AND RWS-SLEEP}
We present additional results for both MWS and RWS, when the recognition model is trained using prior samples.

{

\subsection{STRING CONCEPTS EXPERIMENT}
\centering
\begin{tabular}{l|cccc}
            & $K=2$ & $3$ & $5$ & $10$\\
            \hline
            {RWS} & 87.1 & 86.5 & 85.2 & 85.0\\
            {\RWSSleep } & 88.9 & 87.5 & 86.7 & 85.5 \\
            {\Name } & 84.1 & 83.1 & 82.8 & 82.5\\
            {\NameSleep} & 82.5 & 82.7 & 82.8 & 82.8
        \end{tabular}
        \label{tab:regex_sleep}
    \captionof{table}{Marginal NLL}

\subsection{CELLULAR AUTOMATA EXPERIMENT}

    \begin{tabular}{l|cccc|cccc}
            & \multicolumn{4}{|c}{Easy ($d=3$)}& \multicolumn{4}{|c}{Hard ($d=5$)}\\
            & $K=2$ & $3$ & $5$ & $10$& $K=2$ & $3$ & $5$ & $10$\\
            \hline

            {RWS} & 0.74 & 0.51 & 0.17 & 0.09& 10.27 & 5.55 & 3.44 & 2.43\\
            {\RWSSleep } & 3.39 & 3.64& 2.84 & 3.12  & 23.07 & 20.54 & 18.05 & 15.53\\
            {\Name } & 0.01 & 0.01 & 0.02 & 0.02 & 1.24 & 1.15 & 0.90 & 0.75\\
            {\NameSleep } & 0.01 & 0.01 & 0.01 & 0.01 & 5.98 & 5.12 & 4.80 & 4.24 \\
        \end{tabular}
        \label{tab:automata_distance_easy}
    \captionof{table}{Distance from true model}

}

\clearpage
    
\section{STRING CONCEPTS DATASET}
Our {\CSVConcepts } dataset was collected by crawling public GitHub repositories for files with the \textit{.csv} datatype. The data was then automatically processed to remove columns that contain only numbers, English words longer than 5 characters, or common names, and so that each column contained at least 10 elements. We then drew one random column from each file, while ensuring that no more than three columns were included with the same column header. This allows us to reduce homogeneity (e.g. a large proportion of columns had the header 'state') while preserving some of the true variation (e.g. different formats of `date').
The final dataset contains 5 training examples and 5 test examples for each of 1500 concepts. A sample is included below.

{
\smaller
\verb~`#574c1d', `#603a0d', `#926627', `#9e662d', `#9e8952'~\\
\verb~`(206) 221-2205', `(206) 221-2252', `(206) 393-1513', `(206) 393-1973', `(206) 882-1281'~\\
\verb~`ca', `ja', `rp', `tw', `vm'~\\
\verb~`EH15 2AT', `EH17 8HP', `EH20 9DU', `EH48 4HH', `EH54 6P'~\\
\verb~`Exi0000027', `Exi0000217', `Exi0000242', `Exi0000244', `Exi0000250'~\\
\verb~`GDPA178', `GDPA223', `GDPA289', `GDPA428', `GDPA632'~\\
\verb~`YE Dec 11', `YE Dec 14', `YE Mar 11', `YE Sep 08', `YE Sep 12'~\\
\verb~`$0', `$330,000', `$35,720,000', `$4,505,000', `$42,095,000'~\\
\verb~`ODI # 2672', `ODI # 2750', `ODI # 3294', `ODI # 3372', `ODI # 3439'~\\
\verb~`3000-3999', `4000-4999', `5000-5999', `50000-59999', `NA'~\\
\verb~`soc135', `soc138', `soc144', `soc67', `soc72'~\\
\verb~`9EFLFN31', `FE87SA-9', `LD7B0U27A1', `PPB178', `TL88008'~\\
\verb~`+1 212 255 7065', `+1 212 431 9303', `+1 212 477 1023', `+1 212 693 0588', `+1 212 693 1400'~\\
\verb~`CH2A', `CH64', `CH72', `CH76', `CH79A'~\\
\verb~`20140602-2346-00', `20140603-1148-01', `20140603-1929-04', `O0601014802', `O0603155802'~\\
\verb~`BUS M 277', `BUS M 440', `BUS M 490R F', `BUS M 490R TTh', `BUS M 498'~\\
\verb~`-2.9065552', `-3.193863', `-3.356659', `-4.304764', `-4.5729218'~\\
\verb~`#101', `#4/2/95/2', `#79', `#8/110/3-2', `#94/2'~\\
\verb~"'1322563'", "'151792'", "'2853979'", "'5273420'", "'7470563'"~\\
\verb~`F150009124', `F150009169', `F150009180', `F150009181', `F150009346'~\\
\verb~`BA-CMNPR2', `BCOUNS', `JBGD', `JDAE', `OBSB51413'~\\
\verb~`b_1', `e_1', `g_2', `k_1', `o_1'~\\
\verb~`P.AC.010.999', `P.IH.040.999', `P.IH.240.029', `P.PC.030.000', `P.PC.290.999'~\\
\verb~`-00:16:05.9', `-00:19:52.9', `-00:24:25.0', `-00:33:24.7', `-00:44:02.3'~\\
\verb~`APT', `FUN', `JAK', `KEX', `NAP'~\\
\verb~`SC_L1_O3', `SC_L3_O2', `SC_L5_O3', `SC_L6_O1', `SC_L6_O2'~\\
\verb~`onsen_20', `onsen_44', `onsen_79', `onsen_80', `onsen_86'~\\
\verb~`SDP-00-005', `SDP-02-106', `SDP-04-079', `SDP-06-067', `SDP-08-045'~\\
\verb~`FM0001', `FM0225', `FM2500', `SL0304', `SS0339'~\\
\verb~`BEL', `KOR', `PAR', `POL', `RUS'~\\
\verb~`-0.5423', `-0.702', `0.2023', `0.6124', `0.6757'~\\
\verb~`R0353226', `R0356653', `R0397240', `R0474859', `R0488595'~\\
\verb~`CB', `NA', `SC', `SE', `WE'~\\
\verb~`|S127', `|S23', `|S3', `|S4', `|S5'~\\
\verb~`GO:0008238', `GO:0009259', `GO:0009896', `GO:0034332', `GO:0043270'~\\
\verb~`MN', `MO', `NE', `SD', `WY'~\\
\verb~`F1-D0N343656', `F1-D0N343666', `F1-D0N343669', `F1-D0N343677', `F1-D0N343680'~\\
\verb~`YATN', `YBTR', `YEJI', `YMGT', `YPIR'~\\
\verb~`ABF', `AF', `CBA', `CC', `EAJ'~\\
\verb~`E', `NNE', `NNW', `W', `WSW'~\\
\verb~`A', `F', `G', `Q', `R'~\\
\verb~`bio11', `bio14', `bio16', `bio19', `tmin4'~\\
\verb~`ACT-B06', `MS-ACT-C15', `MS3-08', `MS960931', `MS960961'~
}

\section{CONVERGENCE}
In this section, we describe the limiting behaviour of the MWS algorithm, assuming a sufficiently expressive recognition network $r(z|x)$.

The MWS(-replay) algorithm advocated in the paper permits a strong convergence result as long as the encoding probability of each program $z$ is bounded ( $r(z|x)>\epsilon_z>0$ ) for the top-$M$ programs in the posterior $p(z|x)$. This ensures the optimal $z$s are eventually proposed. While it is simple to enforce such a constraint in the network structure, we have found this unnecessary in practice.

We consider convergence of $r$ and $Q$ in three cases, providing brief proofs below:

\begin{enumerate}
    \item \textbf{Fixed $p(z,x)$; large memory ($M \rightarrow \infty$)}

The encoder $r(z|x)$ converges to the true posterior $p(z|x)$
    \item \textbf{Fixed $p(z,x)$; finite memory}

$r(z|x)$ and $Q_i(z)$ converge to the best $M$-support posterior approximation of $p(z|x)$.

$r(z|x)$ will therefore be accurate if $p(z|x)$ if sufficiently sparse, with $M$ elements of $p(z|x)$ covering much of the probability mass. We believe this condition is often realistic for program induction (including models explored in our paper) and it is assumed by previous work (e.g. Ellis et al. 2018).

    \item \textbf{Learned $p(z,x)$}

p and Q converge to a local optimum of the ELBO, and $r(z|x)$ matches $Q_i(z)$. This is a stronger result than exists for (R)WS: in (R)WS, optima of the ELBO are fixed points, but convergence is not guaranteed due to differing objectives for $p$ and $q$ (although this is rarely a problem in practice).
\end{enumerate}

Note that, for the MWS-\textit{fantasy} variant of the algorithm, $r(z|x)$ is trained directly on samples from $p$, and so necessarily converges to the true posterior $p(z|x)$

\textbf{Proofs}
\begin{enumerate}
    \item If $M$ is large enough to contain the full domain of $z$, then $Q_i(z)$ is equivalent to enumerating the full posterior. $r(z|x)$ is trained on samples from $Q_i(z)$.
    \item $Q$ is updated (to minimize $D_{KL}(Q||P)$) in discrete steps proposed by $r$. Assuming that $r$ eventually proposes each of the top-$M$ values in $p(z|x)$, these proposals will be accepted and $Q$ converges to the optimal set of programs (with $Q_i(z) \propto p(z|x)$). Then, $r(z|x)$ learns to match $Q_i(z)$.
    \item $Q$ and $p$ are optimized to the same objective (ELBO), using SGD for $p$ and in monotonic steps for $Q$. The algorithm therefore converges to a local optimum of ($p$, $Q$), and $r$ converges to $Q_i$.
\end{enumerate}

\clearpage
\section{SPARSITY}
In order to provide empirical support to the theoretically founded suggestion that MWS relies on posterior sparsity, we ran an additional experiment building on the synthetic GMM model described in section 4.1. In this experiment, we vary the cluster variance $\sigma^2$ in the dataset, as a way to produce different levels of posterior sparsity. That is, a large $\sigma^2$ produces more overlap between clusters, which leads to less certainty over which cluster an observation came from (less posterior sparsity).

Below, we show the improvement in posterior accuracy gained by using MWS (compared to VIMCO/RWS) for a range of cluster variances $\sigma^2$ and particles K. We find significant improvement from MWS for low $\sigma^2$ ($\sigma^2$=0.03, as in Figure 4) which diminishes as $\sigma^2$ increases.

\begin{table}[ht]
    \centering
    \begin{tabular}{c|cccc}
         $\sigma^2$ & 0.03 & 0.1 & 0.3 & 1.0 \\
         \hline
         K & \multicolumn{4}{c}{Improvement in $D_{KL}$ by MWS}\\
2   & 10.26  & 3.35  & 0.99  & 0.07\\
5   & 10.21  & 3.57  & 0.73  & -0.06\\
10  & 6.62   & 3.13  & 1.35  & 0.77\\
20  & 5.42   & 3.25  & 2.24  & 2.13
    \end{tabular}
    \caption{The $D_{KL}$ improvement is defined as: $\textrm{min}\Big(
        D_{KL}(Q_\textrm{vimco}||p), \quad D_{KL}(Q_\textrm{rws}||p)\Big)
        - D_{KL}(Q_\textrm{mws}||p)$}
    \label{tab:my_label}
\end{table}

\subsection{MAP INFERENCE}
Note, at the lower limit of $M=1$ (which corresponds to $K=2$ in the table above), MWS may be seen as simply an algorithm for learning with amortized MAP inference.

This task has also been approached with Deep Directed Generative Autoencoders \citep{ozair2014deep}, which are closely related to sparse autoencoders, and may also be seen as a special case of VAEs in which the recognition network $q(z)$ is deterministic.

\section{HANDWRITTEN CHARACTERS MODEL}
\label{appendix-chars}
We use an LSTM for both the prior $p(z)$ and the recognition model $r(z|x)$, where the output is a sequence of (strokeid, on/off) tuples. In the case of alphabet generalisation, both the prior and recognition model are conditioned on the alphabet index.

Each stroke $j$ in the stroke bank is parametrised by its total displacement $\phi_j^x, \phi_j^y$, and parameters $w$ and $u$ to control the stroke width and sharpness. For the likelihood $p_\phi(x|z)$ we use a differentiable renderer to draw strokes on a canvas: each pixel in the image is set to a value of
$$\textrm{max}(0, e^{{\phi_j^u}^2 ({\phi_j^w}^2 - d)})$$
where $d$ is the shortest distance from that pixel to the stroke. During rendering, all strokes are placed end-to-end onto the canvas and we then take the output to be the logits of a Bernoulli likelihood.

We include an additional parameter $\phi_j^a$ to each stroke, which corresponds to the \textit{arc angle}, where $\phi_j^x, \phi_j^y$ still correspond to the straight-line distance from the start point to the end point. This angle ranges from $-2\pi$ to $2\pi$ with no discontinuity, corresponding to the two possible orientations of the circle (clockwise or anticlockwise).

We learn our model using ADAM, using 10 characters per alphabet as training data.

To evaluate the log marginal likelihood $\log p(x)$, we fix the generative model, and refine the recognition model by training for additional $20$k iterations (batch size = $50$) with $K = 200$.
We use this refined recognition model to estimate $\log p(x)$ using the IWAE bound with $5$k samples.
We run this procedure with RWS, VIMCO and Sleep-fantasy training and report the best $\log p(x)$.

\clearpage

\begin{figure*}[b]
    \centering
    \includegraphics[width=\textwidth]{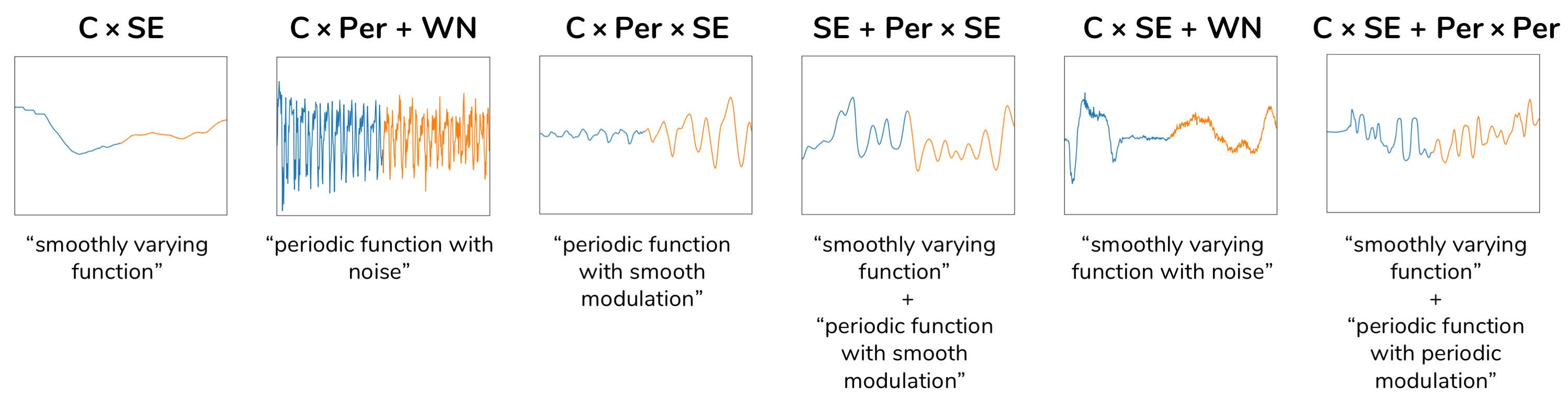}
    \caption{Learning to model the UCR time series dataset with Gaussian Processes, by inferring a latent kernel $z$ for each observed timeseries. Blue (left) is a 256-timepoint observation $x_i$, and orange (right) is a sampled extrapolation using the inferred kernel $z \sim Q_i$ (symbolic representation above, natural language representation below). During learning, we use $Q_i$ as a memory for the discrete structure of kernels, but use a Variational Bayes inner loop to marginalise out a kernel's continuous variables when evaluation of $p(z,x)$ is required.}
    \label{fig:kernels}
\end{figure*}

\section{TIMESERIES DATA}
\label{appendix-timeseries}
As a preliminary experiment, we applied MWS to the task of finding explainable models for time-series data. We draw inspiration from \citet{duvenaud2013structure}, who frame this problem as Gaussian process (GP) kernel learning. They describe a grammar for building kernels compositionally, and demonstrate that inference in this grammar can produce highly interpretable and generalisable descriptions of the structure in a time series.

We follow a similar approach, but depart by learning a set of GP kernels jointly for each in timeseries in a dataset, rather than individually. We start with time series data provided by the UCR Time Series Classification Archive. This dataset contains 1-dimensional times series data from a variety of sources (such as household electricity usage, apparent brightness of stars, and seismometer readings). In this work, we use 1000 time series randomly drawn from this archive, and normalise each to zero mean and unit variance.

For our model, we define the following simple grammar over kernels:
\begin{align}\label{kernelgrammar}
K \rightarrow K + K \mid K * K \mid \textrm{WN} \mid \textrm{SE} \mid \textrm{Per} \mid \textrm{C}\textrm{,\hspace{1em} where}
\end{align}
\begin{itemize}
    \itemsep0em
    \item WN is the \textit{White Noise} kernel, $K(x_1,x_2) = \sigma^2 \mathbb{I}_{x_1=x_2}$
    \item SE is the \textit{Squared Exponential} kernel, $K(x_1,x_2) = \exp(-(x_1-x_2)^2/2l^2)$
    \item Per is a \textit{Periodic} kernel, $K(x_1,x_2) = \exp(-2\sin^2(\pi|x_1-x_2|/p)/l^2)$
    \item C is a \textit{Constant}, $c$
\end{itemize}

We wish to learn a prior distribution over both the symbolic structure of a kernel and its continuous variables ($\sigma$, $l$, etc.). To represent the structure of the kernel as $z$, we use a symbolic kernel `expression': a string over the characters 
\begin{align*}
    \{(, ), +, *, \textrm{WN}, \textrm{SE}, \textrm{Per}, \textrm{C}\}
\end{align*}
We define an LSTM prior $p_\theta(z)$ over these kernel expressions, alongside parametric prior distributions over continuous latent variables ($p_{\theta_\sigma}(\sigma), p_{\theta_l}(l), \ldots$). As in previous work, exact evaluation of the marginal likelihood $p(x|z)$ of a kernel expression $z$ is intractable and so requires an approximation. For this we use a simple variational inference scheme which cycles through coordinate updates to each continuous latent variable (up to 100 steps), and estimates a lowerbound on $p(x|z)$ using 10 samples from the variational distribution.

Examples of latent programs discovered by our model are displayed in Figure \ref{fig:kernels}. These programs describe meaningful compositional structure in the time series data, and can also be used to make highly plausible extrapolations.

\clearpage